\documentclass[manuscript,screen,nonacm,preprint]{acmart}

\renewcommand\footnotetextcopyrightpermission[1]{} 
\usepackage{caption}
\usepackage{verbatim}
\usepackage{mathtools}

\usepackage{subcaption}
\usepackage{xcolor}
\usepackage{float} 

\usepackage[vlined,ruled]{algorithm2e}

\def\prob{\mathbb{P}}

\def\natural{\mathbb{N}}

\usepackage{amsmath,graphicx,epsfig,color,amsfonts}

\graphicspath{{./fig/}}

\def\prob{\mathbb{P}}

\def\natural{\mathbb{N}}

\newcommand{\supscr}[2]{#1^{\textup{#2}}}

\newcommand{\seqdef}[2]{\{#1\}_{#2}}

\newcommand\oprocendsymbol{\hbox{$\square$}}
\newcommand\oprocend{\relax\ifmmode\else\unskip\hfill\fi\oprocendsymbol}

\def \mc {\mathcal}

\renewcommand{\theenumi}{(\roman{enumi}}

\AtBeginDocument{%
  }

\begin{document}


\title{When To Seek Help: Trust-Aware Assistance-Seeking in Human-Supervised Autonomy}

\author{Dong Hae Mangalindan}
\email{mangalin@msu.edu}
\affiliation{%
  \institution{Michigan State University}
  \city{East Lansing}
  \state{Michigan}
  \country{USA}
}

\author{Ericka Rovira}
\email{ericka.rovira@westpoint.edu}
\affiliation{%
  \institution{United States Military Academy}
  \city{West Point}
  \state{New York}
  \country{USA}}

\author{Vaibhav Srivastava}
\email{vaibhav@egr.msu.edu}
\affiliation{%
  \institution{Michigan State University}
  \city{East Lansing}
  \state{Michigan}
  \country{USA}
}

\begin{abstract}

Our goal is to model and experimentally assess trust evolution to predict future beliefs and behaviors of human-robot teams in dynamic environments. Research suggests that maintaining an appropriate level of trust among team members in a human-robot team is vital for team success. Indeed, research suggests that trust is a multidimensional and latent entity that relates to past experiences and future actions in a complex manner. Employing a human-robot collaborative task, we model human behavior using Input-Output Hidden Markov Model (IOHMM) and design an optimal assistance-seeking strategy for the robot using a Partially Observable Markov Decision Process (POMDP) framework. In this task, the human partner supervises an autonomous mobile manipulator collecting objects in an environment. The supervisor's responsibility is to ensure that the robot safely executes its task.  The robot can either choose to attempt to autonomously collect the object or seek human assistance. The human supervisor actively monitors the robot's activities, offering assistance upon request, and intervening if they perceive that the robot may fail. In this setting, human trust is the hidden state, and the primary objective is to optimize team performance. We conduct two sets of human-robot interaction experiments. The data from the first experiment are used to estimate POMDP parameters, which are used to compute an optimal assistance-seeking policy evaluated in the second experiment. The estimated POMDP parameters reveal that, for most participants, human intervention is more probable when trust is low, particularly in high-complexity tasks. Notably, our estimates suggest that the robot's action of asking for human assistance in high-complexity tasks can positively impact human trust. Our experimental results show that the proposed trust-aware policy results in a better performance as compared to an optimal trust-agnostic policy. By comparing model estimates of human trust, obtained using only behavioral data, with the collected self-reported trust values, we show that model estimates are isomorphic to self-reported responses.

\end{abstract}




\maketitle

\section{Introduction}
Autonomous systems are becoming increasingly prevalent and are being deployed and used in various sectors such as healthcare, agriculture, military, and transportation.
Despite the increasing capabilities of autonomous systems, these systems still require human supervision and assistance when working in complex and uncertain environments\cite{
Selma2017,akash2020human,peterssupervisor2015, gupta2024structural,Boguslavskii2025}. However, the increased capabilities of the robots are leading to novel landscapes of human-robot interaction (HRI)  wherein the robots are deployed in roles requiring complex social behavior with human partners.

Human trust in the robotic system is a key factor to be considered for effective collaboration between humans and robots in complex and uncertain environments. While trust has been traditionally studied in the context of collaborative human teams~\cite{mayer1995integrative}, its influence in heterogeneous human-robot teaming is gaining prominence with the increased capabilities and deployment of robotic systems. 
Lee and See \cite{lee2004trust} define trust in an autonomous agent as ``the attitude that an agent will help achieve an individual’s goals in a situation characterized by uncertainty and vulnerability''. Wagner and Arkin~\cite{wagner2011recognizing} generalize this definition to define trust as ``belief held by the trustor that the trustee will act in a manner that mitigates the trustor's risk in a situation in which the trustor has put its outcome at risk.'' Trust in a robotic system is not only influenced by its physical design but also by the autonomous decisions and actions executed by the robot.

While having low trust is undesirable as it leads to the disuse of a system, having high trust can also be undesirable as it can lead to misuse or abuse of the system~\cite{hoff2015trust,hancock2011meta,parasuraman1997humans}.
Therefore, for an effective design of human-robot teams, the calibration of the human partner's trust in the robot's abilities and capabilities is necessary. To this end, it is vital to understand the influence of various robot-related, environment-related, and human-related factors on the dynamic evolution of trust.

The classical approach to studying the impact of robot reliability, transparency, and human workload on human trust treats trust as a static parameter~\cite{hoff2015trust}. 
However, recent studies have focused on understanding the dynamic evolution of human-automation trust, focusing on factors such as the quality of robot performance, transparency of the robot's operation, and the human agent's attitude and characteristics, such as forgetfulness and self-confidence, on the trust dynamics~\cite{lee1992trust, lee1994trust, manzey2012human,desai2013impact, akash2017dynamic, yang2017evaluating, Wang2015DynamicRS, chen2020trust,akash2020human,williams2023computational}.
Hancock \emph{et al.}~\cite{hancock2011meta} present a comprehensive review of various factors affecting human trust and categorize them into three major groups: robot-related, human-related, and environment-related factors. 
They discuss that human trust is most influenced by robot-related factors, which include robot reliability, performance, design, and abilities. This is followed by the environment-related factors, which include task complexity, type, and physical environment. Among the three main categories, human-related factors such as demographics, expertise, and personality have the least influence on trust in human-robot interaction.

For efficient collaboration with humans, a robot teammate should be aware of the human's intent and plan, which are influenced by human trust in the robot. To this end, a trust-modulated human behavior model is essential for the design of robot policies. The human behavior model enables the inference of variables such as human trust and the prediction of human action, intent, preference, and plan, which can be used for planning robot actions for more efficient team performance.
Additionally, since robot actions may influence human trust in a few interaction instances, such a model should capture the dynamic evolution of trust with the above factors as input. 
Existing models in the literature can be categorized into deterministic and stochastic models.

Deterministic linear models use a linear time-invariant dynamical system, where the states are the current trust level, cumulative trust level, and expectation bias~\cite{akash2017dynamic, yang2017evaluating, Wang2015DynamicRS}. The input to these linear models includes human experience, which is determined by the performance and reliability of the robot. Variants of these models have been introduced to capture mutual trust between humans and robots \cite{rahman2018mutual, Wang2014}. These models have been used in several contexts, including exploring the impact of information transparency in reconnaissance missions \cite{yang2017evaluating}, investigating human reliance on driving assistance systems \cite{akash2017dynamic}, examining subtask allocation in collaborative assembly tasks \cite{rahman2018mutual}, and addressing scheduling challenges in teleoperation for underwater robotic navigation \cite{Wang2014}.

Stochastic models of trust often take one of two forms. Stochastic linear models for the evolution of trust~\cite{azevedo2021real} rely on linear time-invariant dynamical systems with Gaussian measurement noise. In contrast, other stochastic models treat trust as a categorical latent variable and estimate its distribution based on observable factors like robot performance and human actions \cite{OPTIMO, liao2024trust}. An example of this class of models is the Input-Output Hidden Markov Model (IOHMM) with categorical states, which have been previously used to model human trust~\cite{akash2020human,DBLP:journals/corr/abs-2009-11890,williams2023computational}. These models are particularly suitable for capturing interpretable internal states (e.g., “High Trust” or “Low Trust”). In addition, unlike linear Gaussian systems, which typically rely on richer or continuous signals such as self-reported trust on Likert scales \cite{chen2020trust}, gaze duration, or attention time \cite{azevedo2021real}, or the presence of a secondary task, the IOHMM 
provide flexibility for handling discrete observations such as human decisions (e.g., Rely or Intervene, Yes or No, Agree or Disagree) in both training and prediction. 
A comprehensive survey of different trust models is presented in \cite{wang2023human}. To optimize robot decision-making under uncertainty about internal human states, Partially Observable Markov Decision Process (POMDP) framework has been employed with trust as latent state influenced by robot actions, human responses, and contextual factors \cite{chen2020trust, zahedi2023trust, akash2020human, 10.5555/2906831.2906852, DBLP:journals/corr/abs-2009-11890}.
It has been used in applications including scheduling in human-supervised robotics to reduce human interventions~\cite{chen2020trust} and monitoring load~\cite{zahedi2023trust}, as well as generating optimal recommendations and explanations based on inferred human state~\cite{akash2020human,10.5555/2906831.2906852}. In this work, we combine IOHMM-based trust modeling with a POMDP-based decision-making framework.

In addition to a dynamic trust model, real-time estimation of trust is vital for the design of trust-based policies that are robust to modeling uncertainties.
Traditionally, human trust is assessed through self-reporting, such as surveys \cite{schaefer2016measuring,lee1992trust,muir1989operators}. 
However, a limitation of this approach is its impracticality for frequent trust survey completion during the experiment.
Psycho-physiological measurements\cite{gebru2022review}
have also been used to infer the trust state; however, the sensing devices required to collect these measurements, e.g., an EEG head cap, limit their applicability. 
For a more detailed review of trust measurements, refer to \cite{gebru2022review}.
These challenges have prompted researchers to rely purely on context and behavioral data, such as human actions, to estimate trust. 
This approach has been implemented and tested in several contexts such as reconnaissance mission \cite{akash2020human}, autonomous driving~\cite{azevedo2021real,DBLP:journals/corr/abs-2009-11890}, and supervision\cite{OPTIMO}.
We focus on similar methodology for estimating human trust: subjective  measures and behavioral data.

In this paper, in the context of human-supervised object collection tasks, we design and validate trust-aware policies for an autonomous system seeking assistance from a human supervisor. In contrast to existing works, our focus is on a scenario where the robot can seek assistance from the human supervisor, preventing a loss of trust and possibly repairing it. 
The major contributions of this work are the following. First, we develop a novel experiment to explore how a robot's requests for assistance influence human trust. Second, we use an Input-Output Hidden Markov Model (IOHMM)~\cite{536317} to capture the evolution of the hidden human trust state. In our formulation, the inputs are the robot's action, environmental complexity, and the outcome or experience in the previous trial, characterized by the success or failure of the robot in accomplishing its goals, and the output is the human action.
The present research sought to develop a model of trust in human-robot teams. It differs from previous modeling efforts in 3 different ways. First we investigated trust evolution in a human-robot collaborative task where we evaluated the human’s intervention in the robot’s actions and we sought to understand how human trust is affected if the robot asked for assistance.
Using data from multiple participants across several trials, we estimate the parameters of an IOHMM model to explore how assistance-seeking, environmental complexity and prior experience influence human trust and action. 
The trust estimates with the estimated IOHMM parameters are compared with self-reported survey data.
We then combine the IOHMM-based trust modeling with a POMDP-based decision-making framework to design an optimal assistance-seeking policy, comparing its effectiveness with a baseline trust-agnostic policy in further experiments.
While IOHMMs and POMDPs have previously been used to model trust and design robot policy, to the best of our knowledge, they have not been applied in the context of assistance-seeking behavior within human-robot collaboration. Our formulation captures how a robot's choices to seek or avoid help dynamically influence human trust over time.

An early version of this work~\cite{DHM-ER-VS:22e} was presented at the 2023 American Control Conference. In addition to the results in~\cite{DHM-ER-VS:22e}, this paper reports results from experiments with a significantly larger number of participants, reports uncertainty in the model fits, grounds the computational model by establishing an isomorphism between model predictions and survey responses, and provides a more detailed discussion.

The paper is structured as follows. Section~\ref{sec:int_task} defines the collaborative object collection task. Section~\ref{sec:Trust-Aware} reviews key ideas of IOHMM and POMDP and discusses the design of the optimal trust-aware assistance-seeking policy. The estimated human behavioral model and its implications are described in Section~\ref{sec:Model}. Using the human-in-the-loop experiment, the trust-aware and trust-agnostic policies are compared in Section~\ref{sec:Results}, Section~\ref{sec:discussion} provides an elaborated discussion related to our results, we conclude in Section~\ref{sec:conclusion}, and Section~\ref{sec:appendix} presents supplementary materials in the Appendix.

\section{Human-Supervised Robotic Object Collection Experiment}\label{sec:int_task}

\begin{figure*}[t!]
\centering
\includegraphics[width=0.8\linewidth]{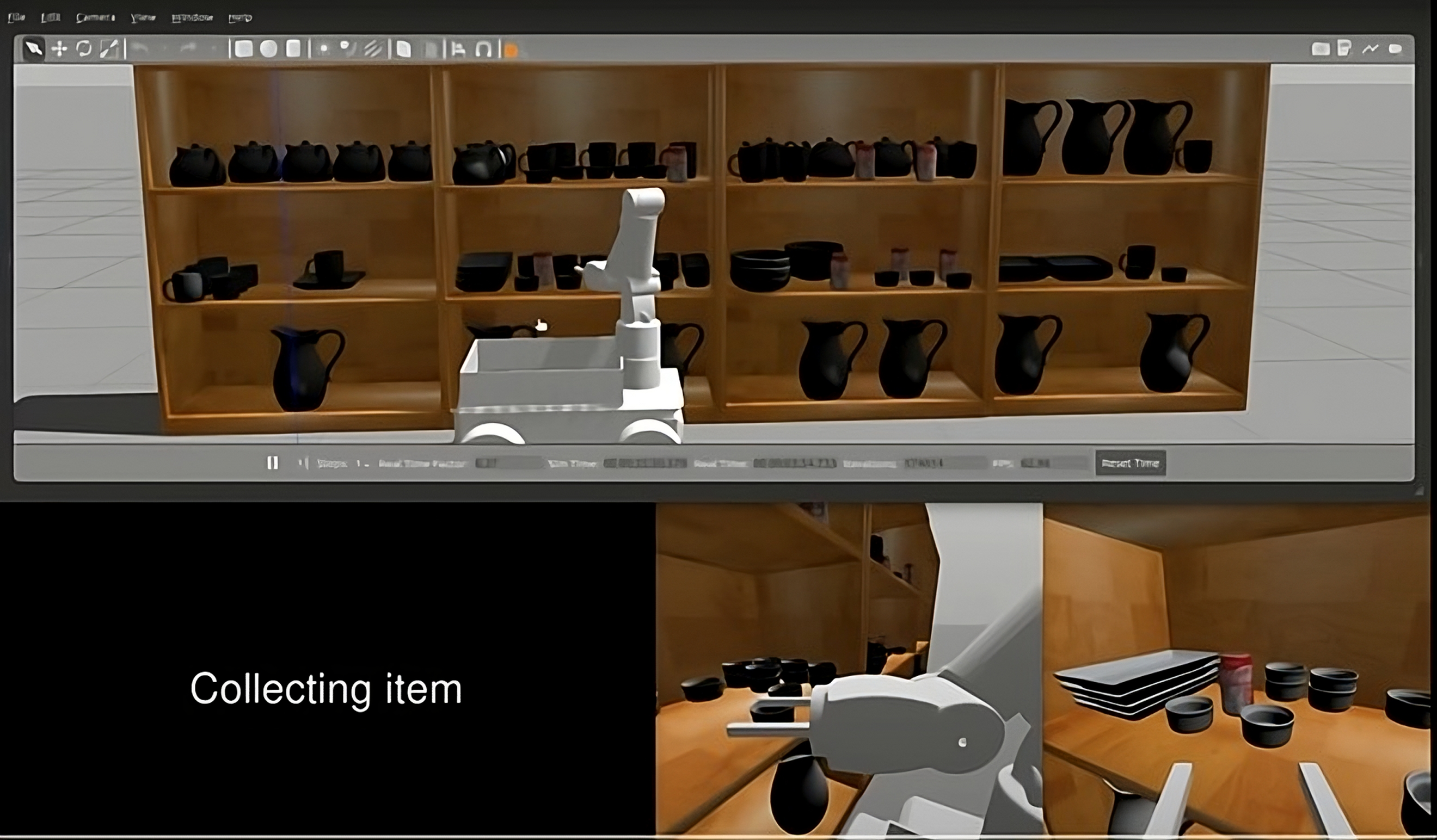} \\
\caption{Experimental Setup. The top figure shows the world view of the environment available to the human supervisor. The bottom left and middle figures show the local view and end-effector view used by humans during teleoperation. The bottom right figure shows the operation status. The setup uses ROS-Gazebo \cite{Gazebo} and resources available in \cite{downs2022google, ROSMoMa}.} 
\Description{}
\label{fig:environment}
\end{figure*}

We focus on a human-supervised object collection task, wherein a human and a robot collaborate to collect objects placed on grocery shelves and deposit them into the bin attached to the robot. The robot used in the experiments is a mobile manipulator equipped with a 4-degree-of-freedom arm and a gripper attached to its end effector. The human supervisor has access to live feeds providing a world view, a local view of the gripper, an end effector view, and a message window to see the task process and also allows the robot to communicate with the human. The human supervisor can teleoperate the arm using a joystick. The virtual environment is set up using ROS-Gazebo. A snapshot of the experiment interface is shown in Fig.~\ref{fig:environment}\footnote{A video of the experiment is available at: https://youtu.be/BhnJHZCVHXQ}.
A trial is identified as each time the robot attempts to collect an object. In each trial, the mobile manipulator has the choice to either attempt to collect the object autonomously ($a^+$) or ask for human assistance ($a^-$). Human assistance involves the human teleoperating the robot and collecting the object. The human supervisor has the choice to either rely on the robot to collect the object ($o^+$) or interrupt the robot before it attempts an autonomous collection and intervene ($o^-$) by collecting the object themselves when there is a perceived risk of the robot failing.

The objects to be collected are placed in two different types of environments: low complexity $C^L$ or high complexity $C^H$. The complexity of the environment is determined based on the presence of obstacles, obstructing the direct path between the manipulator and the object to be collected. 
In our experiment, at the beginning of each trial, the complexity of trial is randomly chosen following an independent and identically distributed (i.i.d.) Bernoulli random variable, with $p_{C^H}$ representing the probability of the complexity being high.

The outcome of each trial can either be a success or failure. Specifically, autonomous object collection is classified as a failure if the robot collides with its surroundings and/or fails to deposit the object in the bin safely. On the other hand, an autonomous collection is deemed a success if the robot safely collects and deposits the object. In our experiment, the success of autonomous collection in low complexity and high complexity, are mutually independent i.i.d Bernoulli variables with success probabilities $\supscr{p_L}{suc}$ and $\supscr{p_H}{suc}$, respectively.

The human experience $E_{t+1}$ is defined by the outcome of trial $t$, and the human and robot's actions. In the context of autonomous operation, the experience is identified as reliable $E^+$ or faulty $E^-$ when the robot succeeds or fails, respectively. In the case of interruption by the human supervisor, the experience is labeled as faulty $E^-$, since the human interrupted and intervened, believing the robot would fail. Given the expectation that the robot should operate unassisted in low-complexity trials, seeking assistance in low complexity is labeled as faulty $E^-$, whereas seeking assistance in high-complexity trials is labeled as reliable $E^+$.

In the experiment, the human supervisor is instructed to maximize the cumulative score/reward of the tasks, where the reward for each trial with the associated robot action, human action, and experience is defined by

\begin{equation}\label{eq:reward}
\mc R^a_{o, E}=
\begin{cases}
+3, & \text{if } (a, o, E) = (a^+, o^+, E^+),\\
+1, & \text{if } a= a^-,\\
0, & \text{if } (a ,o) =  (a^+, o^-),\\
-4, & \text{if } (a, o, E) = (a^+, o^+, E^-).
\end{cases}
\end{equation}

The reward is designed to maximize team performance. Specifically, it maximizes the number of collected objects while minimizing human effort from assistance and intervention. It seeks to reduce unnecessary human interruption and ensure that, by assigning a zero reward for interruption, humans may only interrupt whenever they think the robot will fail the trial.

\section{Trust-aware Optimal Assistance-seeking} \label{sec:Trust-Aware}

\begin{figure}[ht!]
    \vspace{0.03in}
    \centering
    \includegraphics[width=0.8\linewidth, trim = 10 20 10 40 ]{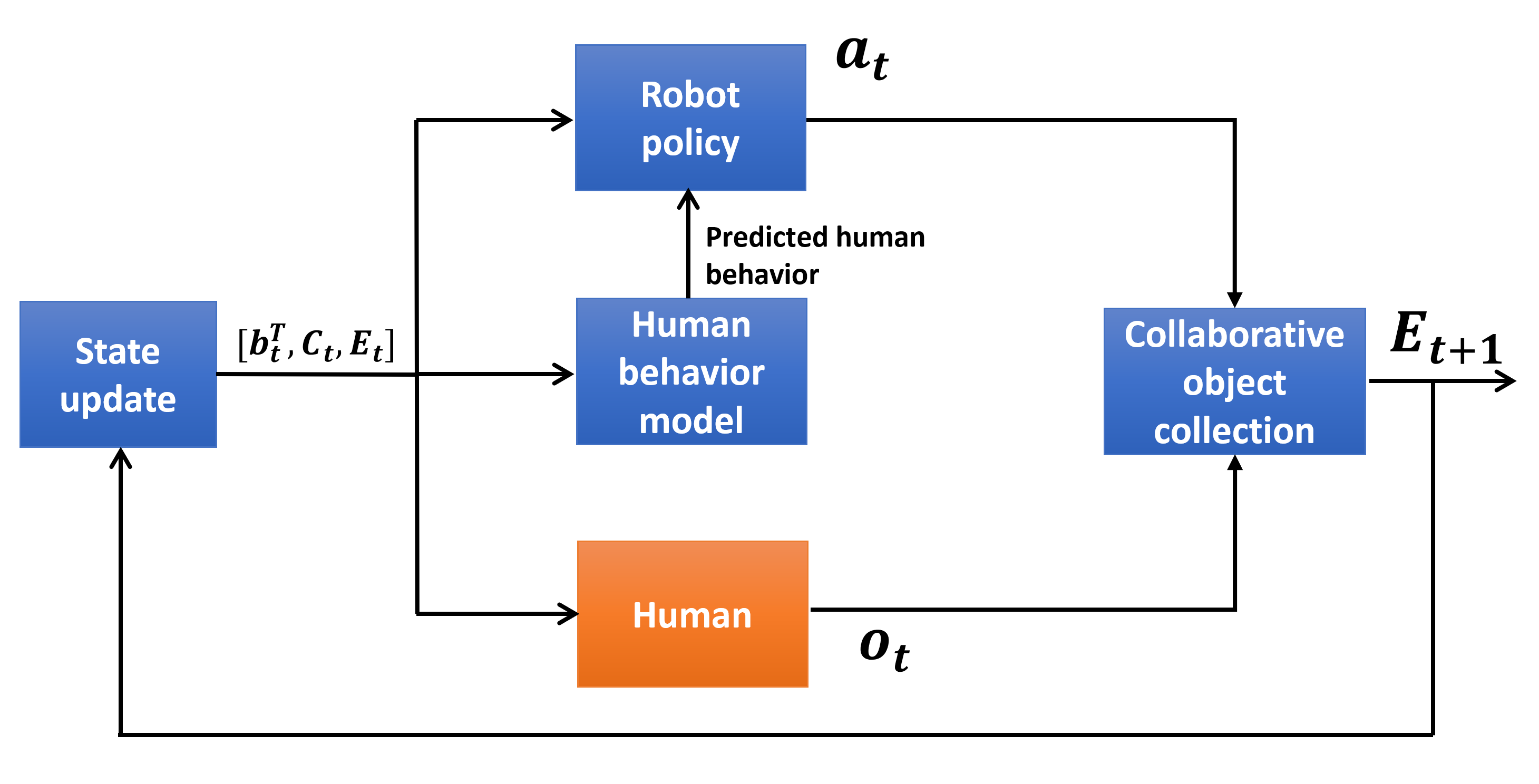}
    
    \caption{Collaborative object collection setup. The human behavioral model with trust as a hidden state yields the likelihood of human intervention and is used by the robot to design an optimal policy in a POMDP setting. }
    \label{fig:block-diagram}
    \vspace{-0.15in}
    \Description{}
\end{figure}

The human-robot system studied in this paper is shown in Fig.~\ref{fig:block-diagram}, and in this section, we describe its different subsystems.

\subsection{Background: IOHMM and POMDP}\label{sec:Background}

Consider  a Markov chain with state $\seqdef{\bar S_t \in \bar{\mc S}}{t \in \natural}$ with an input $\seqdef{u_t \in \mc U}{t \in \natural}$, i.e., $\bar S_{t}$ is a random process such that the probability distribution of $\bar S_{t+1}$ is completely determined by $\bar S_t$ and $u_{t+1}$. Here, $\bar{\mc S}$ and $\mc U$ are some finite sets. In an IOHMM \cite{536317}, 
the state $\bar S_t$ is hidden and not directly observed, and only an output $\seqdef{y_t \in \mc Y}{t \in \natural}$ is measured, which is a realization from an unknown probability distribution $\prob(y_t|\bar S_t,u_t)$. Here, $\mc Y$ is some finite set. The objective of the IOHMM is to infer the hidden state $\bar S_t$ and learn its state transition matrix and $\prob(\bar S_{t+1}|\bar S_t,u_{t+1})$ from  $y_t$ and $u_t$.

A POMDP \cite{krishnamurthy2016partially} extends IOHMM by having some of the inputs controllable, while others are assumed Markovian and included in the state space. Consider the POMDP with augmented state $\seqdef{S_t \in \mc S}{t\in \natural}$, controllable action sequence $\seqdef{a_t \in \mathcal{A}}{t \in \natural}$, and observation sequence $\seqdef{o_t\in \mathcal{O}}{t \in \natural}$. Along with the state set $\mc S$, action set $\mc A$, and observation set $\mc O$, the POMDP 7-tuple is completed by state transition probabilities $\mathbb{P}(S_{t+1}|S_t,a_t)$, observation probabilities $\mathbb{P}(o_{t}|S_{t},a_t)$, a reward function $\mathcal{R}(S_t,a_t)$, and a discount factor $\gamma$. Note that state $S_t$ includes hidden state $\bar S_t$ and non-controlled inputs of the IOHMM. Additionally, $o_{t}$ includes perfect measurements of the non-controlled inputs and output $y_t$ of the IOHMM.

\subsection{Trust-modulated Human Behavior Model}
\label{subsection:HBM}

The human behavior model defines the trust evolution and the action selection choice of the human, defining the probability of the human relying on the robot ($o^+$) whenever the robot does not seek assistance ($a^+$). It is assumed that the human's decision to rely on the robot when it is attempting to collect autonomously is dependent on the complexity of the trial $C_t$, and a hidden state which we refer to as the human trust $T_t$. We assume that the hidden trust state can be high $T^H$ or low $T^L$. 

\begin{figure}[ht!]
\includegraphics[width=0.7\linewidth]{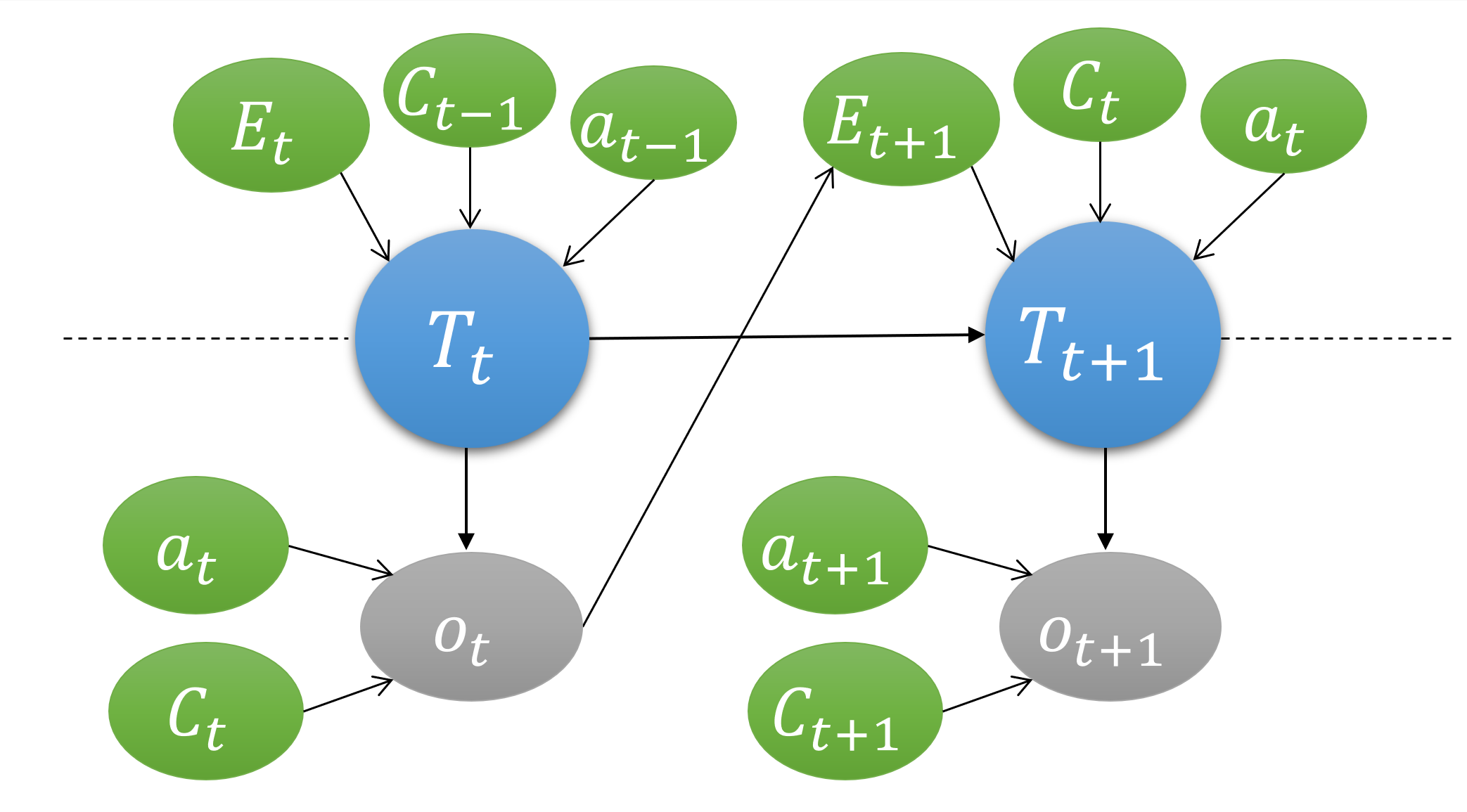}
\caption{IOHMM-based dynamic human behavioral model, where the human action $o_t$ is modulated by the hidden trust dynamics $T_t$}
\Description{}
\label{fig:ext_IOHMM}
\vspace{-0.18in}
\end{figure}

Building upon existing models in the literature \cite{akash2017dynamic}, we let the trust level $T_{t+1}$ at the end of trial $t$ as being influenced by the trust $T_t$ at the beginning of the trial, and the experience $E_{t+1}$ at the end of trial $t$ caused by the robot action $a_t$ at the complexity $C_t$, and the human action $o_t$, during trial $t$. Fig.~\ref{fig:ext_IOHMM} shows the IOHMM-based dynamic human behavior model, which includes the interconnection and influences of the hidden state, inputs, and observable human actions. Training the IOHMM using data sequences results in finding the probability values for trust dynamics $\prob(T_{t+1}|T_t, E_{t+1}, C_t, a_t)$ and human action selection $\prob(o_t|T_t, a_t, C_t)$

\subsection{POMDP-based Optimal Assistance-Seeking Policy}
We formulate the design of an optimal assistance-seeking policy as a POMDP with the following 
states, actions, and observations. The state of the POMDP is defined as $S_t = (T_t, E_t, C_t) \in \{T^L, T^H\} \times \{E^-, E^+\} \times \{C^L, C^H\}$, where $T_t$ is a hidden state and $(E_t, C_t)$ are observed states; the robot actions $a_t \in \{a^+, a^-\}$. The observations $o_t \in \{o^+, o^-\}$, are the observed human actions. 

\noindent
\textbf{State transition matrices and observation probabilities.}
Using the IOHMM structure in Fig.~\ref{fig:ext_IOHMM} and the fact that in each trial, the complexity $C_t$ is i.i.d., the joint state transition probability can be written as
    \begin{align*}
        \prob(S_{t+1}|S_t, a_t)=\prob(T_{t+1}|E_{t+1},T_t,C_t,a_t) \prob(C_{t+1})\prob(E_{t+1}|T_t,C_t,a_t),
    \end{align*}
where the state transition probability of human trust $\prob(T_{t+1}|E_{t+1},T_t,C_t,a_t)$ is obtained from the learned human behavioral model, the probability of complexity being high $\prob(C_{t+1}= C^H)= p_{c^H}$ by design, and the experience transition probability $\prob(E_{t+1}|T_t, C_t,a_t)$ is computed, taking into account the human trust state and action, as
\begin{align*}
\prob(E_{t+1}|T_t,C_t,a_t) =\sum_{o_t\in \mc O}\prob(E_{t+1}|o_t,C_t,a_t)  \prob(o_t |T_t,C_t,a_t)
\end{align*}
where $\prob(o_t |T_t,C_t,a_t)$ is known from the trained human behavior model, and
\begin{align*}
\prob(E_{t+1} = E^+ & |o_t,C_t,a_t) 
= \begin{cases}
\supscr{p_L}{suc}, & \text{if } (a_t, C_t,o_t) = (a^+,C^L, o^+),\\
\supscr{p_H}{suc}, & \text{if } (a_t, C_t,o_t) = (a^+,C^H, o^+), \\
0, &\text{if } (a_t, C_t, o_t) = (a^+, *, o^-), \\
0,  & \text{if } (a_t, C_t, o_t)=  (a^-, C^L,*), \\
1,  & \text{if } (a_t, C_t, o_t)=  (a^-, C^H,*). 
\end{cases}    
\end{align*}

\noindent
\textbf{Reward function and discount factor.} Using the reward defined in~\eqref{eq:reward}, the reward function for the POMDP needs to take into account the probabilities of different human actions given the current state. The reward function for the POMDP is calculated as
\begin{align} \label{eq:pomdp-reward}
\begin{split}
R(S,a^+) &=  \mc R^{a^+}_{o^+, E^+}
\prob(E^+|o^+,C,a^+)\prob(o^+|T,C,a^+)
 +\mc R^{a^+}_{o^+, E^-}\prob(E^-|o^+,C,a^+)\prob(o^+|T,C,a^+)
+ \mc R^{a^+}_{o^-, E^+}\prob(o^-|T,C,a^+),\\
R(S,a^-)&= \mc R^{a^-}_{*, *}, 
\end{split}
\end{align}
where, for brevity, we have represented events by their outcomes; for example, we have represented the event $\{E=E^+\}$ by $E^+$. The discount factor $\gamma$ is chosen as $0.99$.

\subsection{Belief MDP Reformulation of the POMDP}
\label{subsec:belief-MDP}
To calculate and design optimal policy for the POMDP, we reformulate the POMDP as an equivalent MDP with belief states $S^b= (b_t^T, E_t, C_t)$, where $b_t^T = \prob(T_t =T^H| E_{t},C_{t-1},a_{t-1},b_{t-1}^T)$. 
The action set and discount factor remain consistent with the POMDP formulation. The belief update can be rewritten as follows:
\begin{align}\label{eq:belief-update}
b_{t+1}^T
=\prob(T_{t+1}=T^H|T_t=T^H,E_{t+1},C_t,a_t)b_{t}^T + \prob(T_{t+1}=T^H|T_t=T^L,E_{t+1},C_t,a_t)(1-b_{t}^T). 
\end{align}
Similarly, $\prob(E_{t+1}|b^T_t,C_t,a_t)$ is computed by calculating the expected value of $\prob(E_{t+1}|T_t,C_t,a_t)$ over $T_t$. The reward function for the belief MDP is computed by computing the expected value of $R(S_t,a)$ in~\eqref{eq:pomdp-reward} over $T_t$. For computational purposes, we discretize $b^T_t$ and use update~\eqref{eq:belief-update} to compute its state transition matrix.

\section{Estimated Human Behavioral Model}
\label{sec:Model}

To estimate the human behavioral and trust dynamics model parameters, a human subject experiment is conducted\footnote{ The human behavioral experiments were approved under Michigan State University Institutional Review Board Study ID 9452.}. 
A total of 39 participants were recruited, comprising 17 males and 22 females, all of whom were freshman or sophomore college students and the majority were not in the engineering program. To elicit human trust, we have informed the participants that they can receive compensation of up to $\$15$, depending on their performance in the experiment. The experiment lasted on average for 1 hour and participants are compensated with an amount of $\$15.00$ for their participation, regardless of their performance. While all participants ultimately received the full $\$15$, this was not disclosed until after the study. This design element was intended to introduce a perceived risk of loss and thereby establish conditions under which trust could highly influence participants’ decisions.
Participants completed 71 trials in total, with 41 trials in low complexity and the remaining 30 trials in high complexity.
When the robot attempts to collect an object autonomously, it has a success probability of $\supscr{p}{suc}_H =0.75$ in high complexity and $\supscr{p}{suc}_L =0.97$ in low complexity environment. These success rates are chosen given the expectation that low-complexity trials are deemed easier as compared with high-complexity trials. 
In each trial, the robot requests human assistance with a probability of $0.10$ in a low-complexity environment and with a probability of $0.33$ in a high-complexity environment.

Before starting the experiment, participants received a description outlining the task, the choices of robot actions, and the supervisor's role. 
Following this, participants watch a video of the robot operating and collecting objects. They are informed that the robot is able to recognize the complexity with high probability and adjust itself to collect the objects safely, and it may occasionally ask for supervisor assistance. They then proceed to a familiarization phase in a practice environment where they are faced with all possible combinations of complexity and robot actions. They are asked to occasionally interrupt and intervene to get familiarized with controlling the manipulator using the joystick.

The collected data in each trial are human action $o$, robot action $a$, complexity $C$, and experience $E$. 
The data collected from all participants are combined and used to estimate the parameters of the human behavioral and trust dynamics model. Each time a shelf is cleared and objects are collected, human participants are asked to fill out a survey related to their trust. It should be noted that the survey responses are not used for learning the parameters of the human behavior model. The survey responses collected are discussed in Section~\ref{subsec:trust_survey}.
To learn the parameters of the IOHMM-based human behavior model, an extended version of the Baum-Welch algorithm is used~\cite{536317}.\footnote{ Our code is available at https://github.com/donghaem14/Learning-algorithm-for-Input-Output-HMM-with-state-and-observation-inputs.} The estimated parameters of the model are presented and discussed in the following section. For brevity and simplification of notation, the events are represented by their outcomes; for example, $E_t=E^+$ is shortened to $E^+_t$. 
Since the data was collected through in-person experiments, we were able to observe the behaviors of the participants and monitor them. Data collected from 6 participants are omitted due to behaviors observed during the experiment such as lack of attention, constantly checking their phone, and inability to follow the basic instructions of the experiment. This leaves 33 data sequences for model parameter estimation.

\begin{figure}[htb!]
\centering
\begin{subfigure}{.33\textwidth}
  \centering
  \includegraphics[width=\linewidth, trim={0cm 0cm 10cm 0cm}, clip]{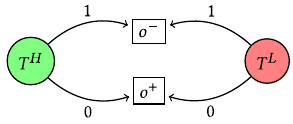}
  \caption{$a=a^-$, $\forall C $}
  \label{fig:EPy}
\end{subfigure}%
\begin{subfigure}{.33\textwidth}
  \centering
  \includegraphics[width=\linewidth, trim={5cm 0cm 5cm 0cm}, clip]{figures/EP_all.pdf}
  \caption{$a=a^+$, $C=C^L$}
  \label{fig:EPnL}
\end{subfigure}%
\begin{subfigure}{.33\textwidth}
  \centering
  \includegraphics[width=\linewidth, trim={10cm 0cm 0cm 0cm}, clip]{figures/EP_all.pdf}
  \caption{$a=a^+$, $C=C^H$}
  \label{fig:EPnH}
\end{subfigure}
\caption {Estimated observation probabilities $\prob(o_t| T, C, a)$}
\Description{}
\label {fig:EP}
\end{figure}

\begin{figure*}[htb!]
\centering
\begin{subfigure}{.33\textwidth}
  \centering
  \includegraphics[width=\linewidth, trim={0cm 0cm 10cm 0cm}, clip]{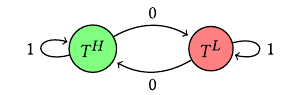}
  \caption{{[$E^+$, $C^L$, $a^+$]}}
  \label{fig:STLRN}
\end{subfigure}%
\begin{subfigure}{.33\textwidth}
  \centering
  \includegraphics[width=\linewidth, trim={5.2cm 0cm 4.8cm 0cm}, clip]{figures/TPL.pdf}
  \caption{{[$E^-$, $C^L$, $a^+$]}}
  \label{fig:STLFN}
\end{subfigure}%
\begin{subfigure}{.33\textwidth}
  \centering
  \includegraphics[width=\linewidth, trim={10cm 0cm 0cm 0cm}, clip]{figures/TPL.pdf}
  \caption{{[$E^-$, $C^L$, $a^-$]}}
  \label{fig:STLA}
\end{subfigure}
\caption {Estimated trust state transition probabilities $\prob(T_{t+1}|T_t, E_{t+1}, C_t, a_t)$ for low environment complexity $C_t=C^L$}
\Description{}
\label {fig:LCST}
\end{figure*}

\begin{figure*}[htb!]
\centering
\begin{subfigure}{.33\textwidth}
  \centering
  \includegraphics[width=\linewidth, trim={0cm 0cm 10cm 0cm}, clip]{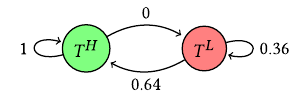}
  \caption{{[$E^+$, $C^H$, $a^+$]}}
  \label{fig:STHRN}
\end{subfigure}%
\begin{subfigure}{.33\textwidth}
  \centering
  \includegraphics[width=\linewidth, trim={5.2cm 0cm 4.8cm 0cm}, clip]{figures/TPH.pdf}
   \caption{{[$E^-$, $C^H$, $a^+$]}}
  \label{fig:STHFN}
\end{subfigure}%
\begin{subfigure}{.33\textwidth}
  \centering
  \includegraphics[width=0.98\linewidth, trim={10.2cm 0cm 0cm 0cm}, clip]{figures/TPH.pdf}
  \caption{{[$E^+$, $C^H$, $a^-$]}}
  \label{fig:STHA}
\end{subfigure}
\caption {Estimated trust state transition probabilities $\prob(T_{t+1}|T_t, E_{t+1}, C_t, a_t)$ for high environment complexity $C_t=C^H$}
\label {fig:HCST}
\vspace{-0.15in}
\Description{}
\end{figure*}

\subsection{Initial Trust and Observation Probabilities.} 
The estimated initial probability of trust being high is $\prob(T^H)=0.82$. This estimate suggests that at the beginning of the experiment, humans have a relatively high trust towards the robot.

The observation probabilities determine the probability of reliance and intervention given the trust level of human, robot action, and the complexity of the trial. The estimated observation probabilities $\prob(o_t| T_t, C_t, a_t)$ are shown in Fig.~\ref{fig:EP}.
The probability of intervention when the robot seeks assistance, $a=a^-$, is $\prob(o^-_t| T_t, C_t,a^-_t)=1$ for any $C$ and $T$. This aligns with the fact that when the robot seeks assistance, the human teleoperates regardless of the trust state and environment complexity. 

The estimated observation probabilities $\prob(o^+_t| T_t, C^L_t, a^+_t)$ is shown in Fig.~\ref{fig:EPnL}. In low complexity, when the robot does not ask for human assistance and attempts to collect autonomously, the human relies with probability 1 and 0.97 when their trust is high and low, respectively. Irrespective of the human trust level, there is a higher probability of reliance in low complexity scenarios due to the higher success rate, with the supervisor perceiving lower risk in such situations.

The estimated observation probabilities $\prob(o^+_t| T_t, C^H_t, a^+_t)$ is shown in Fig.~\ref{fig:EPnH}. In high complexity, when the robot does not ask for human assistance and attempts to collect autonomously, the human relies with probability 0.94 and 0.43 when their trust is high and low, respectively. As compared with the parameters obtained in low complexity, the probability of reliance in high complexity is lower in general. 
The high probability of reliance, when their trust is high, is indicative of humans perceiving the robot to be efficient and reliable; while the high probability of interruption, when their trust is low, is indicative of humans perceiving a higher risk and possible failure in high complexity.

\subsection{State Transition Matrices.}

The estimates obtained for trust state transition in the low complexity environment are shown in Fig.~\ref{fig:LCST}.  The estimates for $\prob(T_{t+1}|T_t,E^+_{t+1}, C^L_t, a^+_t)$ can be seen in Fig.~\ref{fig:STLRN}. If the current trust is high, it remains high with a probability of 1; conversely, if the current trust is low, it remains low with a probability of 1. Success in low complexity does not help in repairing trust, showing that success in an easy task has very little to no effect on restoring trust. 

The estimates for trust transition when there's a failure in low complexity, $\prob(T_{t+1}|T_t, E^-_{t+1}, C^L_t, a^+_t)$, are shown in Fig.~\ref{fig:STLFN}. In this scenario, if the trust is currently high, it transitions to low with a probability of 0.71; while if the current trust is low, it remains low with a probability equal to 1. This shows that failure negatively impacts trust, which is expected.

Fig.~\ref{fig:STLA} shows the estimate of state transition, $\prob(T_{t+1}|T_t, E^-_{t+1}, C^L_t, a^-_t)$, when the robot seeks human assistance in low complexity. The trust state does not change after the trial with probability 1.  Thus, seeking assistance in low complexity has minimal effect on trust, with the trust remaining unchanged.

We will now discuss the state transition when operating in high complexity. Fig.~\ref{fig:HCST} shows the estimates of state transition probabilities for high environment complexity.
Fig.~\ref{fig:STHRN} shows the model estimates for a success in autonomous collection $\prob(T_{t+1}|T_t, E^+_{t+1}, C^H_t, a^+_t)$. It can be seen that if the current trust is high, it stays high with probability 1, and the trust can transition from low to high with probability 0.64. Thus, success in high-complexity tasks can repair trust. 

The estimate of $\prob(T_{t+1}|T_t, E^+_{t+1}, C^H_t, a^+_t)$, the case when autonomous collection fails, is shown in Fig.~\ref{fig:STHFN}. Human trust transitions from high to low with probability equal to 0.33 and remains in low with probability 0.88. Thus, as expected, a failure in high complexity has a negative effect on trust. With the reduced autonomous success rate in high complexity, these tasks are high-risk and high-reward. In summary, a failure negatively impacts the trust level, and success in more complex tasks $C^H$ has a higher chance of repairing trust as compared to success in an easier task $C^L$. 

The estimate of $\prob(T_{t+1}|T_t, E^+_{t+1}, C^H_t, a^-_t)$ is shown in Fig.~\ref{fig:STHA}. When the robot seeks assistance in high complexity, the human trust remains high with probability 1. Interestingly, if the current trust is low, the trust transitions from low to high with probability 0.13. Our estimates suggest that asking for assistance in high complexity may increase trust. In instances where human trust is low, they may perceive the robot as less reliable to operate autonomously. Therefore, seeking assistance can improve trust because the human may perceive the robot as being cautious and avoiding potential failure by not attempting to collect autonomously. Seeking assistance in high complexity can help repair trust and prevent unsolicited intervention by the human supervisor.

\subsection{Model Parameter Confidence}

We now investigate uncertainty in the estimated parameters. Since the Baum-Welch algorithm provides a maximum a posteriori estimate, we rely on Laplace approximation \cite{pml2Book} to estimate parameter uncertainty. To this end, we estimate the Hessian matrix $H$ of the log-likelihood $L$ and compute the uncertainty in parameter estimates as $e=\sqrt{\operatorname{diag}(-H)}$, where $\operatorname{diag}$ represents the diagonal elements of the matrix $-H$. The uncertainties in the estimated IOHMM human trust model parameters are shown in Table~\ref{tb:uncert}.

\begin{table}[ht!]
\centering
\caption{Uncertainty in  IOHMM parameters.
}
\begin{tabular}{|l|c|c|} 
\hline
Probabilities & $C^L$ & $C^H$ \\
\hline\hline
$\prob (T^H|T^H,E^+,a^+)$ & 0.04 & 0.04\\
\hline
$\prob (T^H|T^L,E^+,a^+)$ & 0.09 & 0.27\\ 
\hline
$\prob (T^H|T^H,E^-,a^+)$ & 0.19 & 0.07\\ 
\hline
$\prob (T^H|T^L,E^-,a^+)$ & 0.56& 0.30 \\ 
\hline
$\prob (T^H|T^H,a^-)$ & 0.11 & 0.06  \\ 
\hline
$\prob (T^H|T^L,a^-)$ & 0.34 & 0.17\\ 
\hline
$\prob (o^+|T^H,a^-)$ & 0.00 & 0.02\\ 
\hline
$\prob (o^+|T^L,a^-)$ & 0.01 & 0.15\\ 
\hline
\end{tabular}
\label{tb:uncert}
\end{table}

For parameters related to $C^L$, the standard errors are relatively low and accurate up to the first significant digit of the estimate. Specifically, this refers to the human action probability and $\prob (T^H|T,E^+,a^+)$ parameters. However, the errors for other parameters are significantly higher, with the highest having an error of $\pm0.56$. These high-uncertainty parameters are associated with scenarios involving failures and when the robot asks for assistance in low complexity which are attributed to the low frequency of these scenarios during the experiment.

The parameters conditioned on the state being in high trust $T^H$ have low standard errors, with the highest standard error being $\pm0.07$. This suggests that these parameter estimates are accurate up to the first significant digit. In contrast, the standard errors for parameters conditioned on the state being in low trust $T^L$ are relatively high, ranging from $\pm0.15$ to $\pm0.3$. This high uncertainty can possibly be due to participants not getting into low trust state often, as compared to the high trust state.

\subsection {Trust as a Hidden State}
\label{subsec:trust_survey}

We now investigate if it is justified to refer to the hidden state in our model as ``human trust". During the experiment, after each subblock, participants completed a survey \cite{lee1992trust} in which they responded, on an 11-point scale, to the following question ``To what extent do you trust the automation in this scenario?'', shown in Appendix~\ref{appendix:subblock_survey}. This question was intended to capture participants’ level of trust based on their ongoing experience while performing the task.
This survey is collected each time the robot finishes collecting objects from a shelf, before moving to another shelf. 

Using the model parameters, we calculate the probability of being in a particular state at each time step via the belief update. 
Given the assumption that trust evolution follows a Markov process, the probability of trust being high is an aggregated representation of the trust state based on previous interactions. We compare the probability of trust being high at the end of each shelf-clearing event to the corresponding survey responses

\begin{figure*}
\centering
\begin{minipage}{.5\textwidth}
  \centering
    \includegraphics[width=\linewidth]{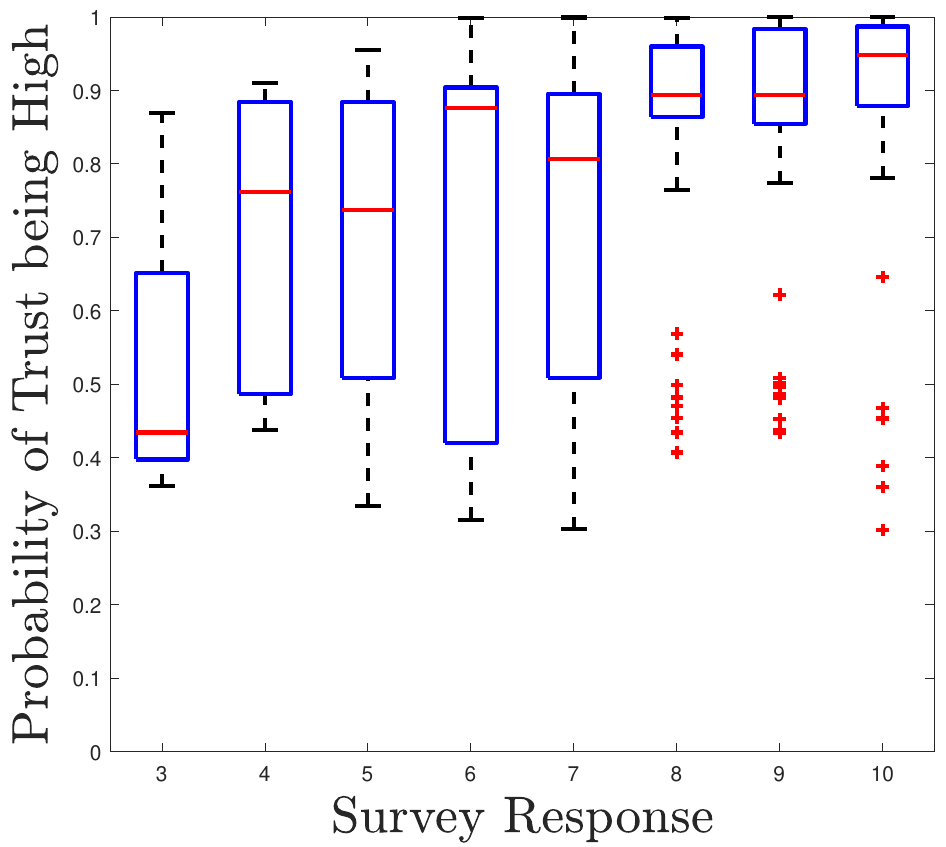}
    \caption{Comparison of trust estimate and survey response }
    \label{fig:subblock_survey}
\end{minipage}%
\begin{minipage}{.5\textwidth}
  \centering
   \vspace{0.4cm}
    \includegraphics[width=\linewidth]{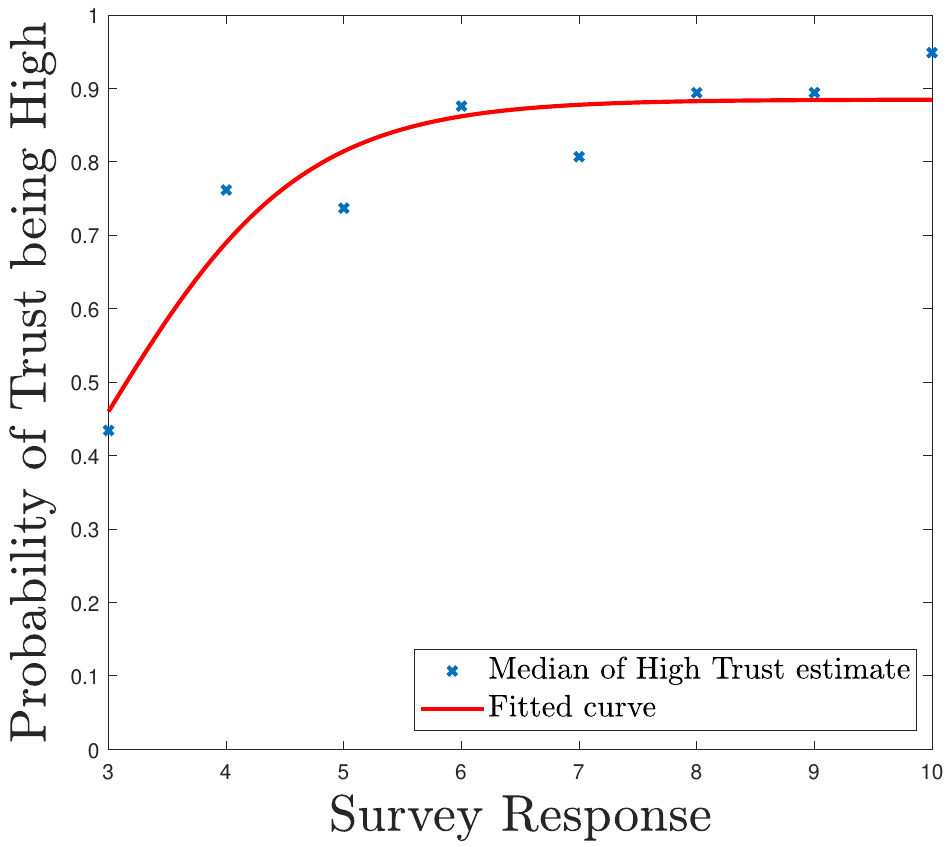}
    \caption{Plot of the fitted curve $P^T(r^T)$ and  median of high trust estimates with respect to trust survey response $r^T$}
    \label{fig:subblock_fitted}
\end{minipage}
\Description{}
\label{fig:test1}
\end{figure*}

The comparison of the probability of trust being high computed using the estimated model, and survey responses is shown in Fig.~\ref{fig:subblock_survey}. It can be observed that the median probability of trust being high increases with trust scores collected through the survey. While the mid-level trust scores have higher variance, the higher trust estimates have lower variance in their trust response. 
Lower trust levels (0,1, and 2) did not receive any rating from participants. 
This can indicate that participants are more certain when their trust is high but uncertain about their trust level when it is low. The median of the probabilities of trust being high is used to fit a curve following a logistic function, resulting in

\begin{align*}
P^T(r^T)=\frac{0.8849}{1+e^{-1.184(r^T-2.932)}},
\end{align*}
where the function $P^T(r^T)$ represents the probability of high trust as a function of the human trust response $r^T$.
The curve is fitted using MATLAB's \texttt{fit} function. The goodness of fit has been calculated using fit function. The Sum of Squares Error (SSE) is calculated to be equal to $0.0213$ and the Root Mean Squared Error (RMSE) is equal to $0.0653$.
The plot of $P^T(r^T)$ with the coefficients as compared with the median values of high trust estimate per survey response can be seen in Fig.~\ref{fig:subblock_fitted}. The results suggest that the hidden state in our model is isomorphic to self-reported trust scores, and the probability of trust being high is a sigmoidal function of the self-reported trust values. Thus, there exists a one-to-one mapping between trust estimates and self-reported survey responses, which suggests that the hidden state in our model is representative of human trust.

\subsection {Trust and Human Reliance}

\begin{figure*}[ht!]
\centering
\begin{minipage}{.5\textwidth}
  \centering
    \includegraphics[width=0.97\linewidth]{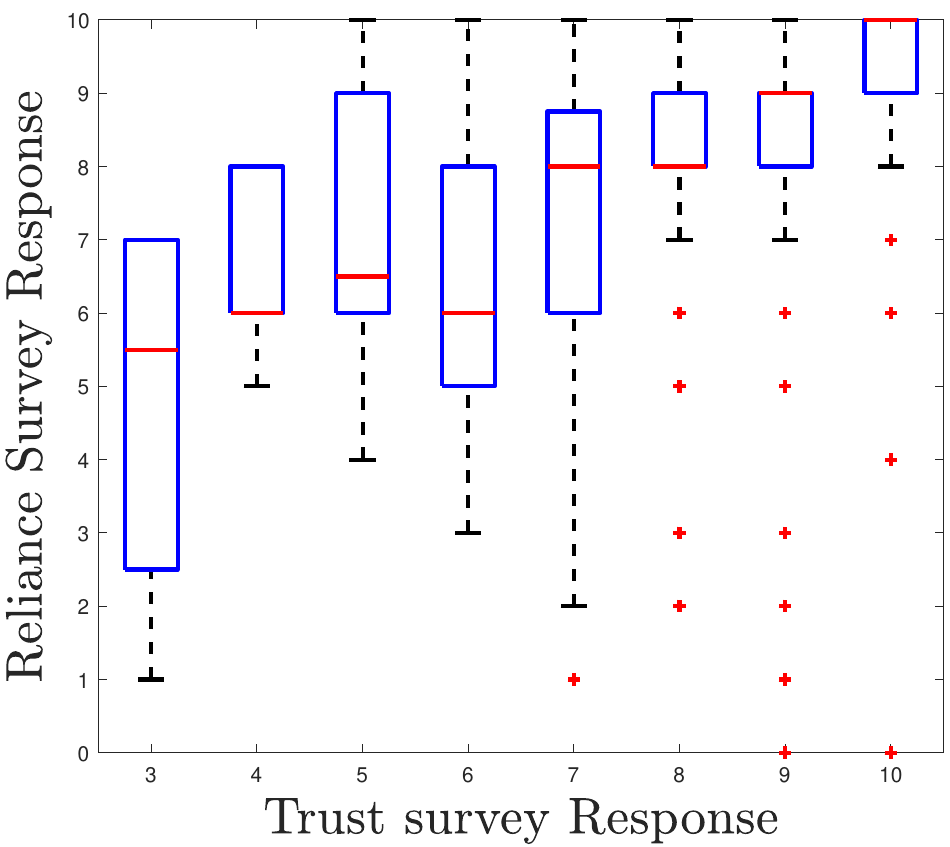}
    \caption{Comparison of trust and reliance survey response.}
    \label{fig:TrustvRel}
\end{minipage}%
\begin{minipage}{.5\textwidth}
  \centering
   \vspace{0.4cm}
    \includegraphics[width=\linewidth]{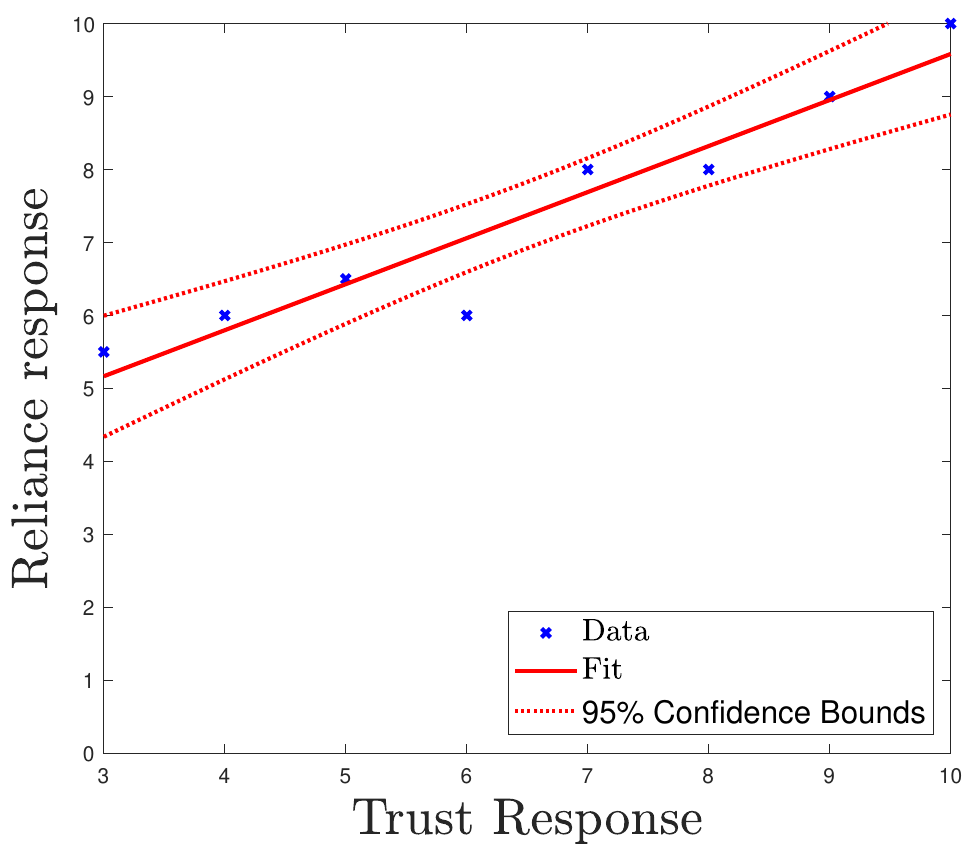}
    \caption{Plot of the linear model which maps the trust response to survey response.}
    \label{fig:TrustvRel_fit}
\end{minipage}
\Description{}
\label{fig:test2}
\end{figure*}

We now analyze and evaluate the relationship between human trust and reliance using survey responses. During the experiment, along with the trust questionnaire discussed in Section~\ref{subsec:trust_survey}, the participants are also asked to fill out a survey after each subblock, in which they answer ``To what extent did you rely on the automation in this scenario'', to which they respond on an 11-point scale, shown in Appendix \ref{appendix:subblock_survey}. The plot of reliance response with respect to the trust response is shown in Fig.~\ref{fig:TrustvRel}. Recall that lower trust levels (0,1, and 2) did not receive any rating from participants.

It is noticed that the median reliance response is gradually increasing with the trust response. Additionally, there is a higher variance in reliance response in lower trust ratings compared to those in higher trust ratings. This can be attributed to human uncertainty regarding their reliance rating when trust ratings are low, in contrast to when trust ratings are high.

It can be observed that the median survey responses can be approximated as a linear function of the trust survey response. The estimated function which maps trust response to reliance response can be approximated by the function
\begin{equation*}
\text{Rel}(r^T)=0.63r^T+3.27,
\end{equation*}
where $\text{Rel}(r^T)$ is the reliance response as a function of trust response $r^T$. The fitted model with the confidence bounds is shown in Fig.~\ref{fig:TrustvRel_fit}. The intercept value of $3.27$ suggests that even in a low trust state, humans may still rely on the robot even in high complexity, as observed in our emission probability estimate.
Additionally, note that our computational model yields only the probability of reliance when the trust is either high or low, and it computes the probability of reliance for a given probability of trust being high by linear interpolation of these reliance probabilities, which is consistent with the linear mapping between reliance response and trust response shown in Fig.~\ref{fig:TrustvRel_fit}.

\section{Validation of Trust-aware Assistance-Seeking Policy}\label{sec:Results}

In this section, we introduce the optimal assistance-seeking policy designed using the estimated POMDP in Section~\ref{sec:Model}. We then validate its efficacy using a second set of human subject experiments where the computed optimal policy is deployed and compared with a baseline policy.

\subsection{Computed Assistance-seeking Policy}

Using the POMDP parameters estimated in Section~\ref{sec:Model}, and the belief-MDP equivalent formulation in Section~\ref{subsec:belief-MDP}, we computed an optimal-assistance seeking policy using the value iteration algorithm with discount factor $\gamma=0.99$.
Even though included in the state, the experience $E_t$ does not influence the policy. This lack of influence might stem from the fact that experience $E_t$ only influences human behavior through their trust $T_t$ (see Fig.~\ref{fig:ext_IOHMM}).
The computed trust-aware optimal policy is to never seek assistance in low-complexity trials. However, for high-complexity trials, the optimal policy transitions from seeking assistance to not seeking assistance as the belief $b^T$ that the human trust is high surpasses a threshold of approximately 0.73:
\begin{align*}
a(t) = \begin{cases}
a^+,  & b^T >0.73, C=C^H, \\
a^-, & \text{otherwise}.
\end{cases}
\end{align*}

The optimal policy focuses on gathering higher rewards by autonomously collecting in high-complexity trials only if the human trust is sufficiently high, avoiding possible interruption by the human supervisor. In the case when human trust is low, it focuses on building trust by seeking assistance and avoiding interruption.

The baseline policy that will be used is a trust-agnostic policy, which does not take the trust state into account. For the trust-agnostic policy, the only attribute of a trial is the environmental complexity. We used the data from the experiment in Section~\ref{sec:Model} to estimate the probability $\prob(o_t=o^-|C_t,a^-)$, which is calculated by taking the empirical estimate in the dataset as
\begin{align*}
\prob(o_t=o^-|C_t,a^+)=\frac{\xi_{(o^-,C_t,a^+)}}{\xi_{(C_t,a^+)}},
\end{align*}
where $\xi_{(o^-,C_t,a^+)}$ is the number of times the human interrupted when the robot action is $a^+$ in $C_t$, and $\xi_{(C_t,a^+)}$ is the number of times the action is $a^+$ in $C_t$.  This is calculated to be $\prob(o_t=o^-|C_t^H,a^+)=0.1848$ in high complexity and $\prob(o_t=o^-|C_t^L,a^+)=0.006$ in low complexity.
The expected reward $\mathbb{E}[R|a^+,C]$ for action $a^+$ for each complexity $C^L$ and $C^H$ is calculated as

\begin{align*}
\mathbb{E}[R|a^-,C] &=\mc R^{a^+}_{o^+, E^+} \prob(o_t=o^+|C_t,a^+) \supscr{p}{suc}_C + R^{a^+}_{o^+, E^-} \prob(o_t=o^+|C_t,a^+) (1-\supscr{p}{suc}_C) 
+ R^{a^+}_{o^-} \prob(o_t=o^-|C_t,a^+),
\end{align*}
where $\prob(o_t=o^+|C_t,a^-)=1-\prob(o_t=o^-|C_t,a^-)$. 

Since the complexity of the environment follows an i.i.d. random variable, for trust-agnostic policy, it suffices to simply calculate the expected reward per action for a given complexity to obtain the optimal action.
With the data collected and our experiment design and parameters, the estimated rewards for attempting to collect autonomously are $\mathbb{E}[R|a^+, C^L]=2.79$ and $\mathbb{E}[R|a^+,C^H]=1.02$. Comparing the expected rewards for each action in the respective complexities, these calculations suggested that the optimal trust-agnostic policy is to never seek assistance irrespective of the environmental complexity, $a(t)=a^+$ for each $t$.

To highlight the importance of including task complexity in the assistance-seeking policy, we calculated a complexity-agnostic optimal policy. Specifically, we calculated an optimal policy that omits task complexity from its formulation. This is calculated by marginalizing over the complexity to obtain the state transition probability, which does not take complexity into account.
The optimal action under this policy is to always collect autonomously, similar to the trust-agnostic policy.

\subsection{Evaluation of the Proposed Policy}
\label{subsec:policy_perf}

\begin{figure}[ht!]
\vspace{-0.1in}
\includegraphics[width=8.4cm]{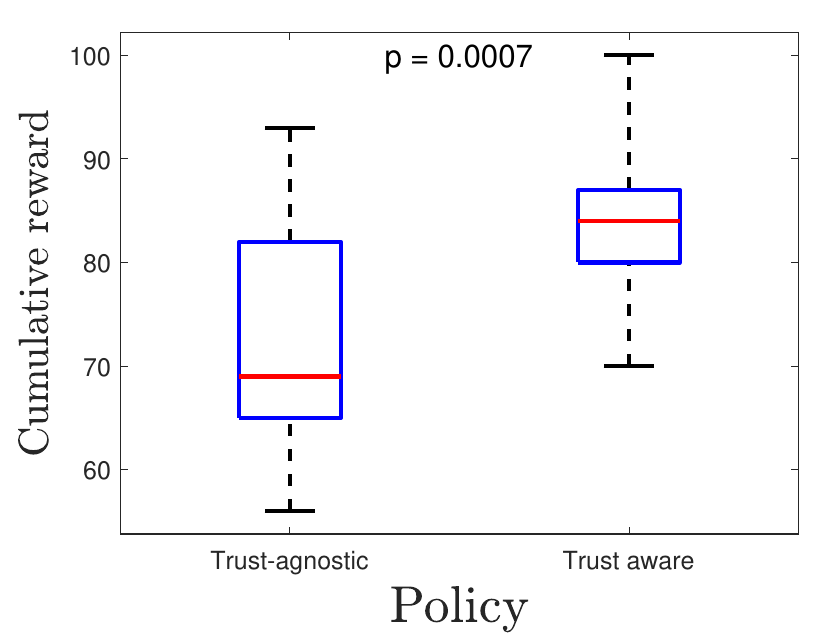}
\caption{Cumulative reward statistics for the trust-agnostic and trust-aware policies. The trust-aware policy outperformed the trust-agnostic policy with a p-value of: 0.0007} 
\label{fig:Score}
\vspace{-0.18in}
\Description{}
\end{figure}

To evaluate the performance of the calculated optimal trust-aware assistance-seeking policy, a second set of experiments with human participants is conducted.
A total of 20 participants were recruited and participated in the experiment, each completing two blocks of experiments in which the robot followed trust-aware and trust-agnostic policies, respectively. The participants recruited are different from those recruited in the first set of experiments.
The order of the blocks was randomly assigned for each participant, with 10 participants starting with the trust-aware policy and the other 10 starting with the trust-agnostic policy. In each block, participants performed 20 low-complexity and 20 high-complexity trials.
Due to the same reason as for the first experiment, data from 2 participants were was omitted, leaving us with 18 data sequences.
The cumulative reward for all participants under both policies is illustrated in Fig.~\ref{fig:Score}. The usage of trust-agnostic policy resulted in a median score of 69, while the trust-aware policy resulted in a median score of 84. The trust-aware policy collected more objects autonomously and prevented more human interruptions as compared with the trust-agnostic policy.
A t-test with $5\%$ significance level shows that trust-aware policy outperforms trust-agnostic policy with a p-value of $p=0.0007$, and is calculated using MATLAB's \texttt{ttest2} function.

We ran additional experiments where participants' compensation is dependent on the team's performance. We informed participants that they will receive a base compensation of $\$10$ and they can receive a bonus based on their aggregate performance/score. The average compensation received by the participants was $\$13.18$. The results are shown in Fig.~\ref{fig:supp_exp_combined} in Appendix~\ref{appendix:supp_exp}. The results are consistent with the above results, i.e., the trust-aware policy outperformed the trust-agnostic policy and the trust-aware policy received a higher trust score from participants.

\subsection{Survey-Based Evaluation of the Trust-Aware Policy}
\label{subsec:policy_trust}

We now investigate if the trust-aware policy improved human trust in addition to improving the team performance as discussed above. In our experiments, participants completed the Trust Perception Scale survey both before and after the experiment. 
The \textit{Trust Perception Scale} is a 40-item survey that aims to measure trust while taking into account the human, robot, and environmental elements of the human-robot interaction. It is designed as a pre- and post-interaction measurement of trust and is used to evaluate changes in human trust \cite{schaefer2016measuring}. The collected responses are the 14 sub-scale of the survey developed in~\cite{schaefer2016measuring}, also shown in Appendix~\ref{appendix:survey_schaefer}.

\begin{figure}[ht!]
\vspace{-0.1in}
\includegraphics[width=8.4cm]{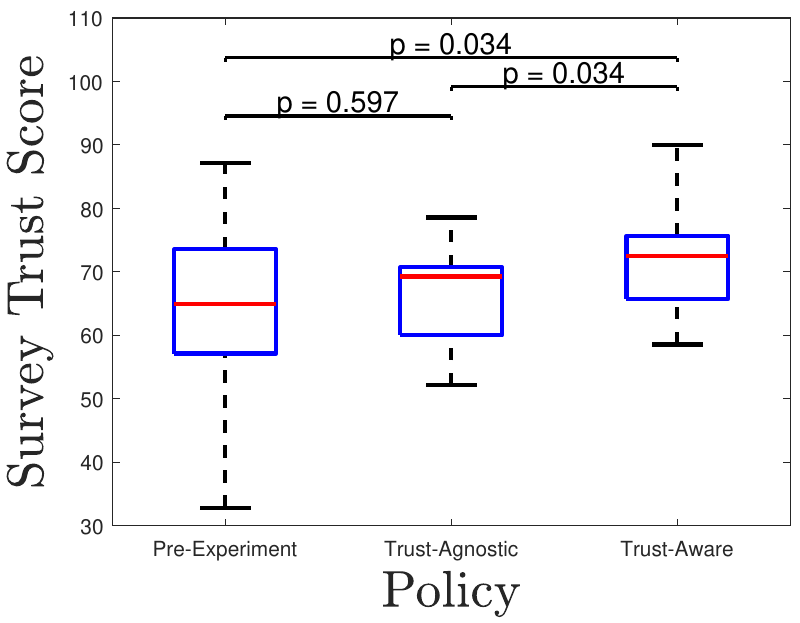}
\caption{Comparison of the collected trust surveys before the experiment and after the experiment with the trust-aware and trust-agnostic policy. } 
\Description{}
\label{fig:Schaefer}
\end{figure}

In total, participants completed this questionnaire three times, the first before the experiment, and the remaining two after completion with each policy. We dub these survey responses as ``pre-experiment", ``post-trust-agnostic", and ``post-trust-aware" responses.
The comparison of the survey scores is shown in Fig.~\ref{fig:Schaefer}. The responses to the 14 survey questions are aggregated, resulting in a trust score.  This process involves first reverse coding the questions that describe negative qualities. The responses are then summed and divided by the total number of questions. More detailed instructions can be found in \cite{schaefer2016measuring}.
The average scores on the collected trust survey are 65.44, 66.39, and 72.14 for pre-experiment, trust-agnostic, and trust-aware, respectively. 

When comparing surveys collected after the experiment, the trust-aware policy (mean score of 72.14) yields a higher trust score relative to the trust-agnostic policy (mean score of 66.39) with a p-value of 0.0342, which is statistically significant. When comparing surveys collected before the experiment (mean score of 64.64) and post-trust-agnostic policy (mean score of 66.39), the p-value calculated is equal to 0.5973, which suggests that trust-agnostic policy has no statistically significant effect on trust. 
A comparison of the surveys collected before the experiment (mean score of 64.64) and after the experiment with the trust-aware policy (mean score of 72.14) suggests that the trust-aware policy improves human trust and the increase is statistically significant with a p-value of 0.0338. We have added a statistical summary of the collected data for this survey, which can be found in Appendix~\ref{appendix:survey_schaefer}. In addition, to support the result, we have also collected trust and reliance survey responses after each subblock, and a statistical summary of the collected data can be seen in Appendix \ref{appendix:subblock_survey}.

\section{Discussion}
\label{sec:discussion}

The concept of trust has been explored across various domains, including both human-human and human-robot interactions. Mayer  \emph{et al.}~\cite{mayer1995integrative} formally define trust between individuals as ``the willingness of a party to be vulnerable to the outcomes of another party based on the expectation that the other will perform a particular action important to the trustor, irrespective of the ability to monitor or control that other party.''.

The definition emphasizes that trust becomes crucial when vulnerability, risk, or uncertainty is involved in an interaction.
With the deployment of robots working in close proximity with human partners in risky and uncertain environments~\cite{takayama2008beyond}, human trust in robots becomes a crucial factor for their adoption.
While human-robot trust shares commonality with human-human trust, the former is typically considered only from the human's perspective regarding the system's performance \cite{baker2018toward}. However, researchers have begun exploring bidirectional trust in human-robot interactions \cite{rahman2018mutual, Wang2014}.

We developed an IOHMM model for trust-modulated human behavior and estimated its parameters using data collected from the experiment.
Along with the definition of trust in automation, Lee and See~\cite{lee2004trust}, provided a conceptual model of trust, describing the process and interaction of factors when taking trust into account.
Our IOHMM model complies with Lee and See's description of how factors interact in the dynamic process when taking trust into account~\cite{lee2004trust}. In our model, the information assimilation is represented by the input to the IOHMM, and the trust evolution is governed by the state transition probabilities. Intention formation is decided by the emission probabilities and human action is represented by the IOHMM output.
Our fitted model suggests that success can increase trust, while failure can decrease it. The reliability of the system is a key factor that shapes human trust in robots \cite{hancock2011meta}. It is well known that positive experiences or successes increase human trust, while negative experiences or failures decrease it \cite{chen2020trust, akash2020human, azevedo2021real}. Additionally, humans tend to rely more on robots when less risk is involved \cite{chen2020trust, akash2020human}. As expected, our model indicates that success in high-complexity tasks has a greater impact on increasing trust compared to success in low-complexity tasks.

We investigated the robot's action of requesting assistance from human supervisors, enabling the robot to transfer control of the current task to the human supervisor. Our model estimates suggest that asking for assistance in high-complexity tasks can help increase trust. 
It has been shown that humans are more trusting of risk-averse robots \cite{bridgwater2020examining}. When human trust is low, their belief that the robot will fail is high, and by seeking assistance, humans perceive the robot as being cautious and avoiding potential failure by not attempting to collect autonomously.

We grounded our model by confirming that its hidden state is isomorphic to self-reported human trust.
Trust model parameters have been estimated using self-reported trust from human participants~\cite{chen2020trust,akash2017dynamic,zahedi2023trust,mangalin2024}. Besides reported trust levels, models have also been estimated using behavioral data, such as action and task performance, from human participants~\cite{azevedo2021real,akash2020human,DBLP:journals/corr/abs-2009-11890,mangalin2025trustrep,williams2023computational}. Traditionally, when human behavior data is used for model parameter estimation, the state of the system is labeled as the human trust but most work does not directly justify this. In our work, we have verified that our trust model's estimate is isomorphic to participants' trust survey responses. With this, we demonstrated that our model accurately captures the dynamics of human trust based solely on behavioral data, without relying on survey responses.

To investigate human trust accurately, a suitable trust-elicitation experiment must be designed to capture trust dynamics effectively.
The most prominent experimental paradigm to study human-robot trust relies on block arrangements, wherein the robot is highly reliable in one block and less reliable in the other~\cite{salem2015would,manzey2012human}.
While these paradigms provide insights into human-robot trust, they do not generate rich enough data to build dynamic models of trust. To address this challenge, we adopted a randomized experimental design wherein the task complexity, influencing robot success rate, is randomly selected in each trial. This approach more naturally excites various levels of human trust and its interaction with different task outcomes, including unexpected and random failures that humans experience when working with robots. Additionally, we believe this experimental paradigm is a closer representation of real-world human-robot interaction scenarios, and is being increasingly adopted in HRI studies~\cite{akash2020human,azevedo2021real,DBLP:journals/corr/abs-2009-11890}

The IOHMM/POMDP-based framework provides a formal structure for inferring human cognitive state and designing optimal policies for stochastic human-robot systems and is being increasingly used in the HRI research community.
Despite their advantages, the data requirement to fit an IOHMM, and the computational requirements to compute an optimal policy for the POMDP often limit the size of the state space, for example, most trust models leveraging this framework use a binary trust state~\cite{wang2023human}. A potential approach to address these issues is to rely on difference equation models~\cite{jonker1999formal} and rely on techniques such as Model Predictive Control~\cite{rawlings2017model} to design the optimal policy~\cite{mangalin2024}. 

In this work, the computed optimal assistance-seeking policy is a threshold-based policy, robot seeks assistance only when its belief that the human trust is low is above a threshold. 
Sufficient conditions have been established for an optimal policy for a POMDP to admit such a structure~\cite{krishnamurthy2016partially, gupta2024structural}. Future research will leverage these results to generic operational characteristics of the robot that result in such policies.

\section{Conclusions and Future Directions}\label{sec:conclusion}

In the context of human-supervised object collection tasks, we designed an optimal assistance-seeking policy for a mobile manipulator. The human behavior model was formulated as an IOHMM, and parameters were estimated using data collected from human subject experiments. The model-based estimates and survey responses were compared, and we showed that the estimated trust states from the model are isomorphic to the self-reported trust ratings. 
We leveraged this model in a POMDP framework to design an optimal assistance-seeking policy that accounts for the human trust state. We showed that the trust-aware policy outperforms a baseline trust-agnostic policy in human-subject experiments. We also showed, by comparing the survey responses of participants, that their trust is higher after working with the robot with a trust-aware policy as compared to working with the robot with a trust-agnostic policy. A possible direction of future research is to schedule low and high-complexity trials for maximizing overall team performance. Other directions of future research include designing trust-aware policies for non-supervisory human-robot collaboration tasks such as active collaboration tasks. Additionally, extending to an adaptive human behavior model to model user-specific variations is another direction. This would allow the robot to adapt more effectively by learning and responding to the unique trust dynamics of each user.

\section*{Acknowledgements}
This work was supported in part by the NSF award ECCS 2024649 and the ONR award N00014-22-1-2813.

\bibliographystyle{acm}
\bibliography{main}
\section{Appendices}
\label{sec:appendix}
This appendix presents the survey questions that were collected, additional experimental results, and a summary of the survey data that complements the findings discussed in the main text. The \textit{Trust Perception Scale}~\cite{schaefer2016measuring} is presented in Section \ref{appendix:survey_schaefer} along with a summary of participant responses. The trust and reliance survey collected throughout the experiment is shown in Section~\ref{appendix:subblock_survey}, accompanied by a statistical summary. The statistical summary of surveys indicates that participants reported higher levels of trust when interacting with the trust-aware policy compared to the trust-agnostic policy.
Additionally, we conducted experiments in which participants’ compensation depended on team performance. The results showing both task performance and human trust are presented in Section~\ref{appendix:supp_exp}.

\subsection{Trust Perception Scale}
\label{appendix:survey_schaefer}

The \textit{Trust Perception Scale} is a 40-item survey that aims to measure trust while taking into account the human, robot, and environmental elements of the human-robot interaction. It is designed as a pre- and post-interaction measurement of trust and is used to evaluate changes in human trust \cite{schaefer2016measuring}. We have collected the 14 sub-scales of the survey:

\begin{center}
\includegraphics[width=8.4cm]{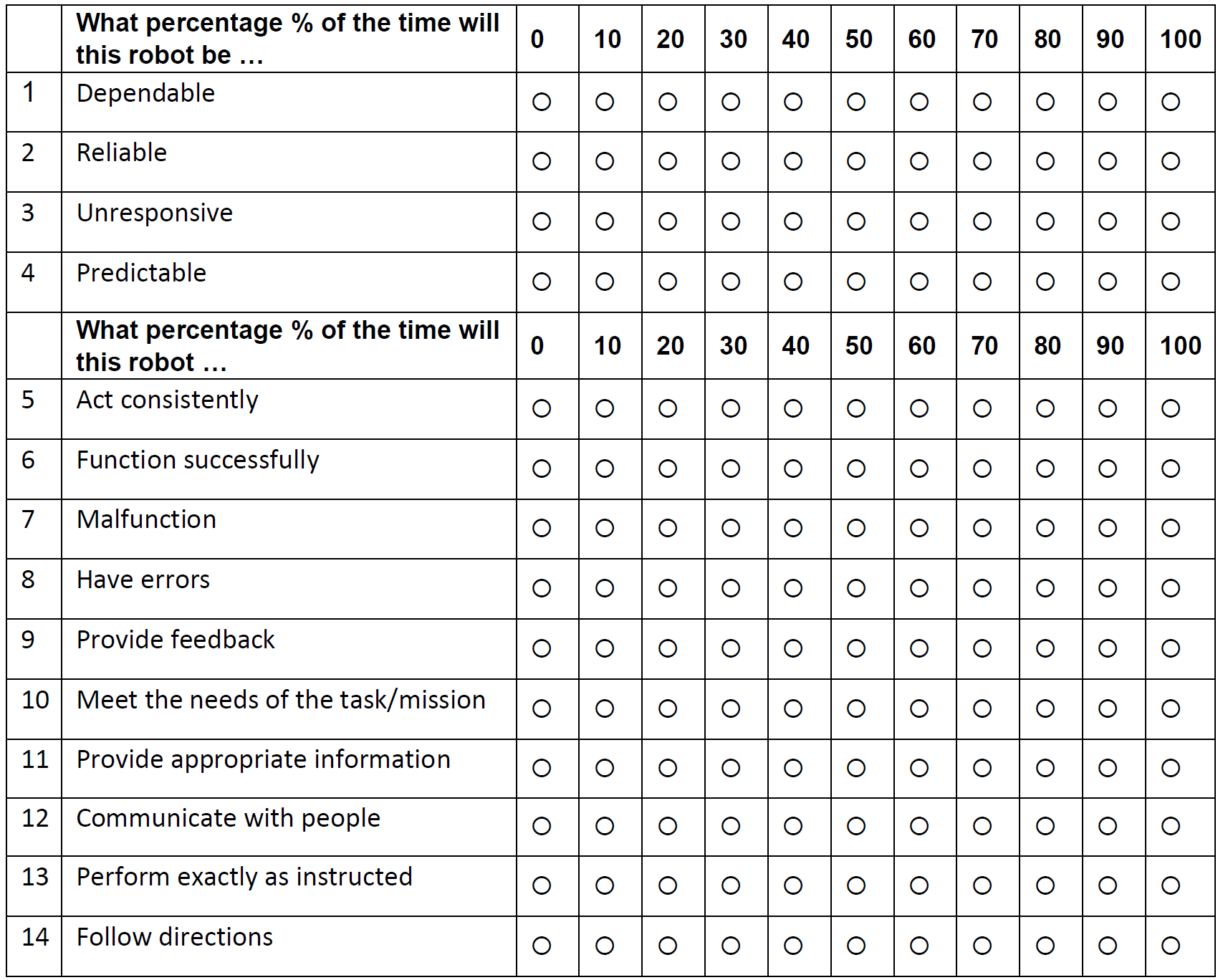}
\end{center} 

We collected survey responses at the beginning of the experiment and after participants interacted with each of the two policies. The summaries of participants’ responses are presented in Table~\ref{tab:combined_survey_stats}, which provides descriptive statistics for all survey items across each experimental condition. The results show that the Trust-Aware policy generally received higher mean ratings for positive characteristics (e.g., dependability, reliability) and lower ratings for negative characteristics (e.g., malfunction) compared to the Trust-Agnostic policy. Overall, the Trust-Aware policy received higher ratings than the Trust-Agnostic policy.

\begin{table}[ht]
\small
\vspace{-1em}
\centering
\
\begin{tabular}{|l|ccc|ccc|ccc|}
\hline
\textbf{Item} 
& \multicolumn{3}{c|}{\textbf{Pre-Experiment}} 
& \multicolumn{3}{c|}{\textbf{Trust-Aware}} 
& \multicolumn{3}{c|}{\textbf{Trust-Agnostic}} \\
& Mean & Median & Variance 
& Mean & Median & Variance 
& Mean & Median & Variance \\
\hline
1  & 64.44 & 70.00 & 226.14 & 76.11 & 80.00 & 154.58 & 69.44 & 70.00 & 217.32 \\
2  & 70.00 & 70.00 & 152.94 & 75.00 & 80.00 & 179.41 & 69.44 & 70.00 & 182.03 \\
3  & 33.89 & 30.00 & 507.52 & 21.11 & 10.00 & 739.87 & 18.33 & 10.00 & 520.59 \\
4  & 67.78 & 70.00 & 430.07 & 77.22 & 80.00 & 127.12 & 70.56 & 70.00 & 229.08 \\
5  & 69.44 & 70.00 & 264.38 & 81.67 & 85.00 & 120.59 & 72.22 & 80.00 & 218.30 \\
6  & 63.89 & 70.00 & 225.16 & 78.89 & 80.00 & 139.87 & 71.67 & 70.00 & 238.24 \\
7  & 37.22 & 25.00 & 844.77 & 22.78 & 10.00 & 562.42 & 27.22 & 25.00 & 515.36 \\
8  & 33.89 & 25.00 & 578.10 & 20.56 & 10.00 & 440.85 & 28.89 & 25.00 & 469.28 \\
9  & 43.33 & 50.00 & 1011.76 & 37.78 & 30.00 & 1041.83 & 32.22 & 20.00 & 1065.36 \\
10 & 66.67 & 70.00 & 247.06 & 80.00 & 80.00 & 141.18 & 73.33 & 75.00 & 188.24 \\
11 & 62.78 & 60.00 & 303.59 & 67.78 & 75.00 & 571.24 & 63.33 & 75.00 & 682.35 \\
12 & 53.89 & 60.00 & 860.46 & 39.44 & 35.00 & 1029.08 & 28.33 & 20.00 & 967.65 \\
13 & 71.11 & 70.00 & 316.34 & 76.67 & 80.00 & 200.00 & 73.33 & 80.00 & 223.53 \\
14 & 76.67 & 80.00 & 400.00 & 83.89 & 90.00 & 154.58 & 80.00 & 80.00 & 329.41 \\
\hline
\textbf{Aggregate} 
& \textbf{64.64} & \textbf{65} & \textbf{138.79} 
& \textbf{72.14} & \textbf{72.50} & \textbf{68.19} 
& \textbf{66.39} & \textbf{69.29} & \textbf{54.17} \\
\hline
\end{tabular}

\caption{ Combined descriptive statistics (mean, median, variance) for each survey item across Pre-Experiment, Trust-Agnostic, and Trust-Aware conditions. It can be seen that, when comparing the mean values for each item, in general, when looking at positive characteristics (i.e., dependability, reliability), the trust-aware policy received a higher mean rating compared to the trust-agnostic policy. Similarly, in items pertaining to negative characteristics (i.e., malfunction), the trust-aware policy received a lower rating compared to the trust-agnostic policy. Aggregating the results, the trust-aware policy received a higher rating compared to the trust-agnostic policy.}
\label{tab:combined_survey_stats}
\end{table}

\subsection{Trust and Reliance Survey}
\label{appendix:subblock_survey}

These surveys are collected after a participant finishes a subblock or shelf. They are used to measure self-reported human trust and reliance throughout the task. These survey questions are based on the seminal work~\cite{lee1992trust,lee1994trust}.

\begin{center}
\includegraphics[width=8.4cm]{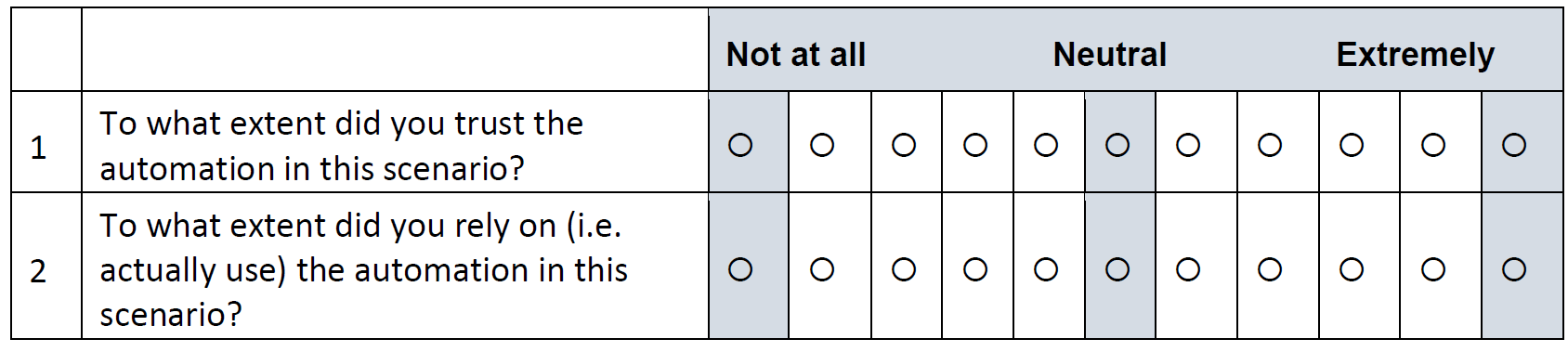}

\end{center}

A statistical summary of the survey responses collected is shown in Table~\ref{tab:trust_reliance_survey}, which compares participants’ ratings for the Trust-Aware and Trust-Agnostic policies. As shown, the Trust-Aware policy received higher trust ratings overall, while the reliance ratings for both policies were approximately the same. These results suggest that participants were generally more trusting of the Trust-Aware policy.

\begin{table}[ht]
\centering
\begin{minipage}{0.45\textwidth}
\centering

\begin{tabular}{lccc}
\textbf{TRUST SURVEY} & Mean & Median & Variance \\
\hline
Trust-agnostic & 6.68 & 7.33 & 4.19\\
Trust-aware    & 7.71 & 7.75 & 2.02 \\
\end{tabular}

\end{minipage}
\hspace{1em} 
\begin{minipage}{0.45\textwidth}
\centering

\begin{tabular}{lccc}
\textbf{RELIANCE SURVEY} & Mean & Median & Variance \\
\hline
Trust-agnostic & 7.66 & 7.83 & 2.92 \\
Trust-aware    & 7.62 & 7.83 & 2.15 \\
\end{tabular}

\end{minipage}
\caption{Trust and reliance survey statistics for trust-aware and trust-agnostic policies. Surveys are collected during the policy evaluation experiment, and we show and compare the statistics for each policy. On average, the trust-aware policy received a higher rating in the trust survey compared to the trust-agnostic policy, while the reliance ratings for both policies were approximately the same. This further suggests that participants were more trusting of the trust-aware policy.}

\label{tab:trust_reliance_survey}
\end{table}

\subsection{Supplemental experiment: Task performance and Trust Score}
\label{appendix:supp_exp}

\begin{figure}[ht]
\centering
\begin{minipage}{0.48\linewidth}
  \centering
  \includegraphics[width=\linewidth]{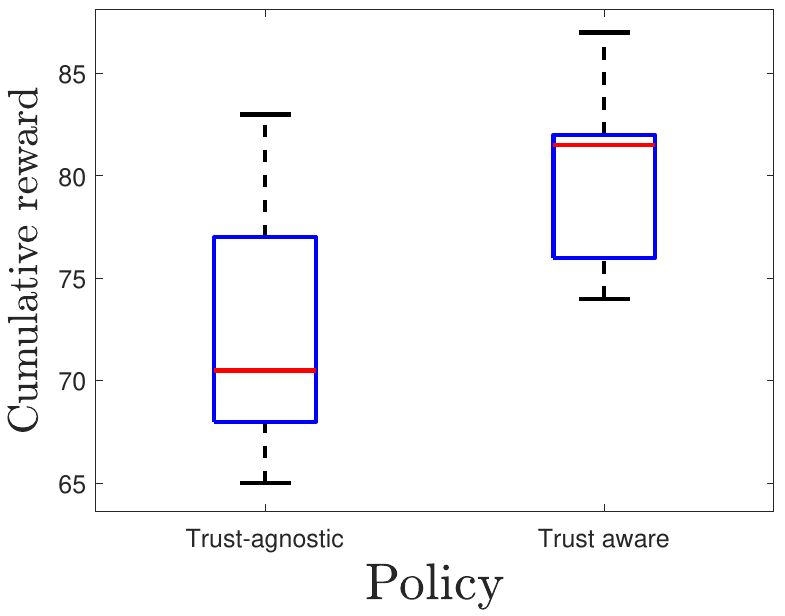}
  \caption*{\textbf{(a)} Cumulative reward statistics for the trust-agnostic and trust-aware policies.  The trust-aware policy outperformed the trust-agnostic policy, consistent with the results discussed in Section~\ref{subsec:policy_perf}.}
\end{minipage}
\hfill
\begin{minipage}{0.48\linewidth}
  \centering
   \vspace{0.1cm}
  \includegraphics[width=\linewidth]{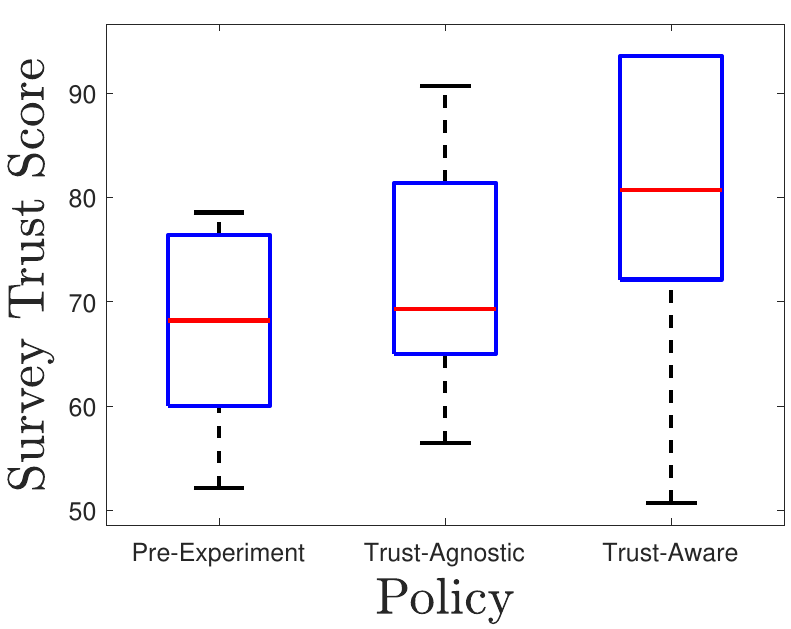}
  \caption*{\textbf{(b)} Comparison of the collected trust surveys before the experiment and after the experiment with the trust-aware and trust-agnostic policy. Results suggest that human trust is higher when working with trust-aware policy, consistent with the results discussed in Section~\ref{subsec:policy_trust}.}
\end{minipage}
\caption{Results from the supplemental experiment on (a) task performance and (b) trust score under performance-based compensation.}
\Description{}
\label{fig:supp_exp_combined}
\end{figure}

We conducted additional experiments with 6 participants, in which their compensation depended on team performance to further incite human trust.  The average compensation received by the participants was $\$13.18$. The results are shown in Fig.~\ref{fig:supp_exp_combined}, and are consistent with the above results, i.e., the trust-aware policy outperformed the trust-agnostic policy, and the trust-agnostic policy received a higher trust score from participants.

\end{document}